# Kinship Verification Based on Cross-Generation Feature Interaction Learning

Guan-Nan Dong, Chi-Man Pun, *Senior Member, IEEE*, Zheng Zhang, *Senior Member, IEEE*

***Abstract*—Kinship verification from facial images has been recognized as an emerging yet challenging technique in many potential computer vision applications. In this paper, we propose a novel cross-generation feature interaction learning (CFIL) framework for robust kinship verification. Particularly, an effective collaborative weighting strategy is constructed to explore the characteristics of cross-generation relations by corporately extracting features of both parents and children image pairs. Specifically, we take parents and children as a whole to extract the expressive local and non-local features. Different from the traditional works measuring similarity by distance, we interpolate the similarity calculations as the interior auxiliary weights into the deep CNN architecture to learn the whole and natural features. These similarity weights not only involve corresponding single points but also excavate the multiple relationships cross points, where local and non-local features are calculated by using these two kinds of distance measurements. Importantly, instead of separately conducting similarity computation and feature extraction, we integrate similarity learning and feature extraction into one unified learning process. The integrated representations deduced from local and non-local features can comprehensively express the informative semantics embedded in images and preserve abundant correlation knowledge from image pairs. Extensive experiments demonstrate the efficiency and superiority of the proposed model compared to some state-of-the-art kinship verification methods.**

***Index Terms*—Kinship verification, face verification, metric learning.**

## I. Introduction

THE human faces not only reveal a wealth of unique individual information but also more familial traits. Due to the particularity of the human face, face-based applications have seen a tremendous surge in interest within multimedia and computer vision communities. Different from the existing research topics on faces, the kin relation verification [1]–[3] aims to distinguish whether there is a biological relationship between the given identities via facial similarity, and has been an emerging research area in computer vision and biometrics verification.

Notably, kinship verification is similar but more challenging compared to face verification. Generally, face verification is based on a single entity, which can identify differences by appearance intuitively. Kinship verification, by contrast, is conducted on multiple entities. As shown in Figure 1, these images are highly similar in face characteristics and muscle movements from genetic diversity. However, they belong to independent individuals. Compared to face verification, kinship verification is more challenging because intra-class variations in kinship are always higher than inter-class variations. More importantly, the feature representations put forward a much higher demand for multiple entities. A large number of kinship verification methods have been proposed and can be roughly grouped into two categories, shallow learning models [4]–[6], and currently popular deep learning models [7]–[9].

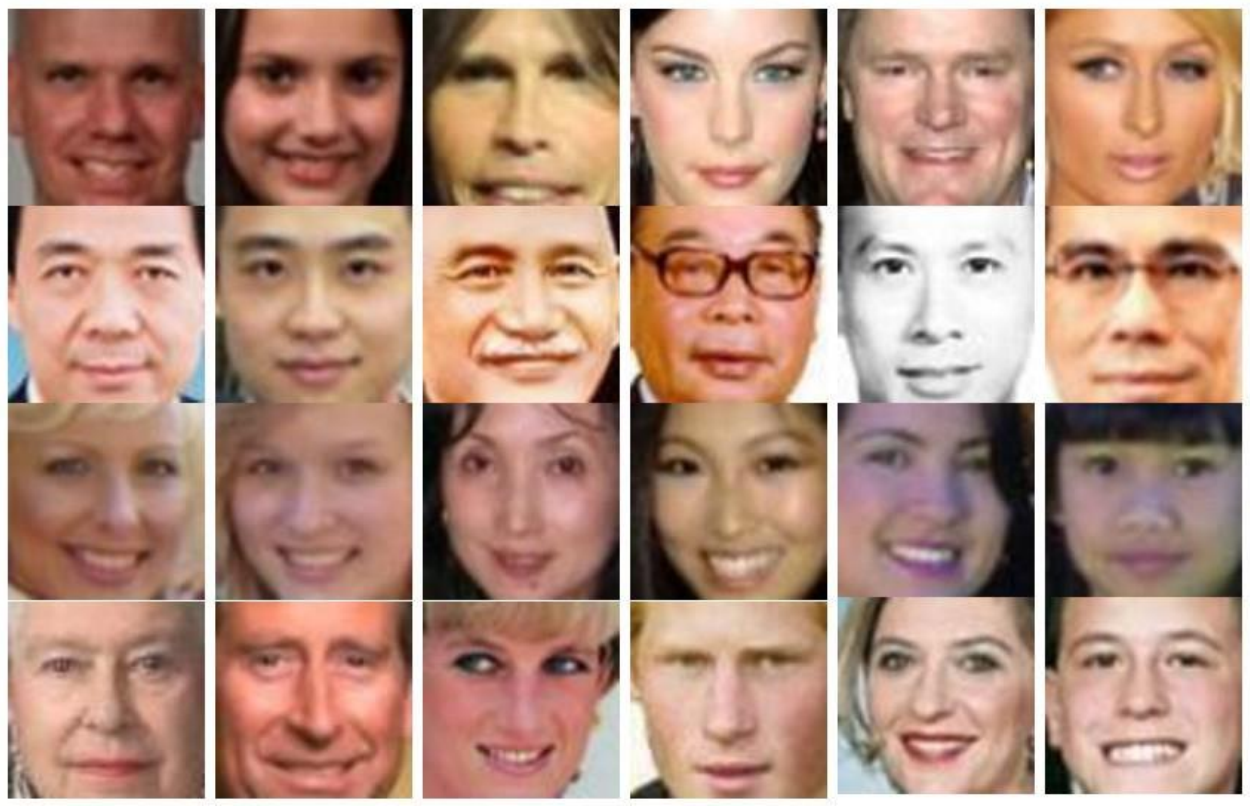

Fig. 1: Positive sample pairs (aligned) from different kin relations. Face images from top to bottom are father-daughter (FD), father-son (FS), mother-daughter (MD) and mother-son (MS), respectively.

The shallow learning models used for kinship verification generally extract the handcrafted features according to different characteristics of faces and then select an appropriate classifier or a distance metric learning (DML) [10] for similarity determination. For example, the spatial pyramid learning method (SPLE) [4] introduces a spatial pyramid feature descriptor for feature extraction and applies the support vector machine (SVM) to kinship verification. The discriminative multi-metric learning (DMML) [6] combines multiple feature extractors, including LBP [11], HOG [12], and multiple metrics to maximize the intra-class coherence and inter-class separabilities of the learned features. The neighborhood repulsed metric learning (NRML) method [5] proposes constructing a distance metric to pull the intra-class samples to similar cluster and push inter-class samples far away from each other, and extracts multiple handcrafted features for multi-view kinship verification based on the SVM classifier. More other related works could be found in a recent survey paper [13]. Although these shallow handcrafted features have been widely used over the past few decades, they lack sufficient representation abilities for big data, leading to

Guan-Nan Dong, Chi-Man Pun, and Zheng Zhang are with the Department of Computer and Information Science, University of Macau, Macau 999078, China. (e-mail: guannandong@outlook.com, cmpun@um.edu.mo, darrenzz219@gmail.com).

inferior performance in complicated computer vision tasks.

Benefiting of the powerful representation capabilities, deep learning has been widely applied in mining the intrinsic underlying structure of face image. Compared to the above shallow methods, deep learning models have demonstrated their stronger adaptability in image feature extraction and classification. The main advantage of deep learning over traditional algorithms is that it does not rely on prior knowledge but learns in an end-to-end manner. For example, a 5-layers shallow CNNs [7] is the first attempt to extract features and verify parent-offspring relations. Moreover, there is another useful strategy, that is, using deep learning model to extract face features and using distance metric learning to measure similarity. For example, the deep kinship verification (DKV) [9] employs an auto-encoder network, followed by a Mahalanobis distance comparison to make intra-class samples more compact.

Although some progress has been achieved, the current best performance kinship verification algorithms typically represent faces with over-complete low-level features and rely heavily on the choice of metric learning algorithms. Specifically, the existing methods pay attention to extracting features in a single image, but fail to uncover the interactive characteristics among different images, which causes their performance to be sensitive to heterogeneous knowledge collected from different sources. Meanwhile, there are no unified and precise representations to collaboratively consider the correlations between cross-generation. In other words, there are no relations and bonds to bridge cross-generation features. Therefore, this separate-based learning paradigm leads to inferior and unstable results due to the mutually exclusive and dependent heterogeneous knowledge. Moreover, the previous works attach great importance to a two-phase independent learning scheme, *i.e.*, feature extraction-independent and similarity metric-independent, which are not embedded into a unified framework to learn synchronously. Notably, they mainly concentrate on the choice of metric learning algorithms, ignoring feature- and heterogeneous-sensitive dependencies for reasonable kinship analysis. These two independent-based learning strategies are hard to build the correlations between cross-generation as well as between feature learning and similarity comparison naturally.

To overcome the above deficiencies, we propose a cross-generation feature interaction learning (CFIL) framework, which jointly considers feature learning and similarity measurement in one unified learning architecture. The outline of the proposed method is graphically illustrated in Figure 2. Specifically, a collaborative weighting strategy for feature extraction is designed to interactively capture the cross-generation relationship of parent-offsprings in a self-attention way. Particularly, the collaborative weighting strategy includes two parts, *i.e.*, the non-local and the local feature extraction module. The non-local feature extraction module is conceived to reflect the semantic relevance of global information, which can provide implicit factors to differentiate the discrete and independent features, while the local feature extraction module is constructed to present the details of the single image to enhance the semantic consistency between image pairs. Notably, when we dexterously integrate the non-local and local features through a collaborative weighting strategy in a mutually reinforced manner, the unified architecture can more naturally express as much semantics as possible and can fully exploit consistency and complementary between images to improve the performance. Furthermore, a unified discriminant learning paradigm is formulated to enhance the dependencies between feature extraction and similarity calculation, and mitigate the effect of possible disturbances in feature learning. Extensive experiments validate the superior performance of the proposed architecture on different benchmark datasets.

The main contributions of this paper are as follows:

- We propose a cross-generation feature interaction learning (CFIL) framework, which combines similarity learning and feature integration to explore cross-generation feature learning under an interactive learning scheme. *To the best of our knowledge*, this is the first attempt to incorporate similarity learning and feature extraction into one unified framework for discriminative kinship verification.
- We obtain the internal relations between parents-children image pairs by introducing a collaborative weighting strategy to capture kinship interactive knowledge. We process the whole facial images to calculate the similarities of different points, which are regarded as interior auxiliary weights embedded in deep CNN representation learning.
- We utilize the distance combination to guide the interactive feature learning and mitigate the enormous divergence of intra-class variations across generations. In this way, more similar information from different perspectives is well-explored to achieve generation-inherent features.
- Extensive experiments conducted on four benchmarks demonstrate the effectiveness of the proposed method for robust kinship verification. Moreover, we further validate our claim that cross-generation feature interaction learning can generate more distinctive feature representations for kinship verification.

The remainder of this paper is organized as follows: Section II introduces the related work. Section III presents the proposed method in detail. Section IV describes our experiments for verifying our claims and rationality analysis. Section V concludes our research and discusses the possible challenges in this area and the scope for further study.

## II. Related Work

In the last decades, extensive efforts have been devoted to kinship verification under a number of learning structures. This section briefly reviews two related research topics: 1) kinship verification, and 2) metric learning.

### A. Kinship Verification

In the human face analysis, children's facial features have a high similarity with those of parents. Therefore, many applications begin to use parental facial images to find missing children, a family genealogy, image searching, annotation, etc. Most kinship analyses are based on learning the most discriminative features for each sample, and some key characteristics

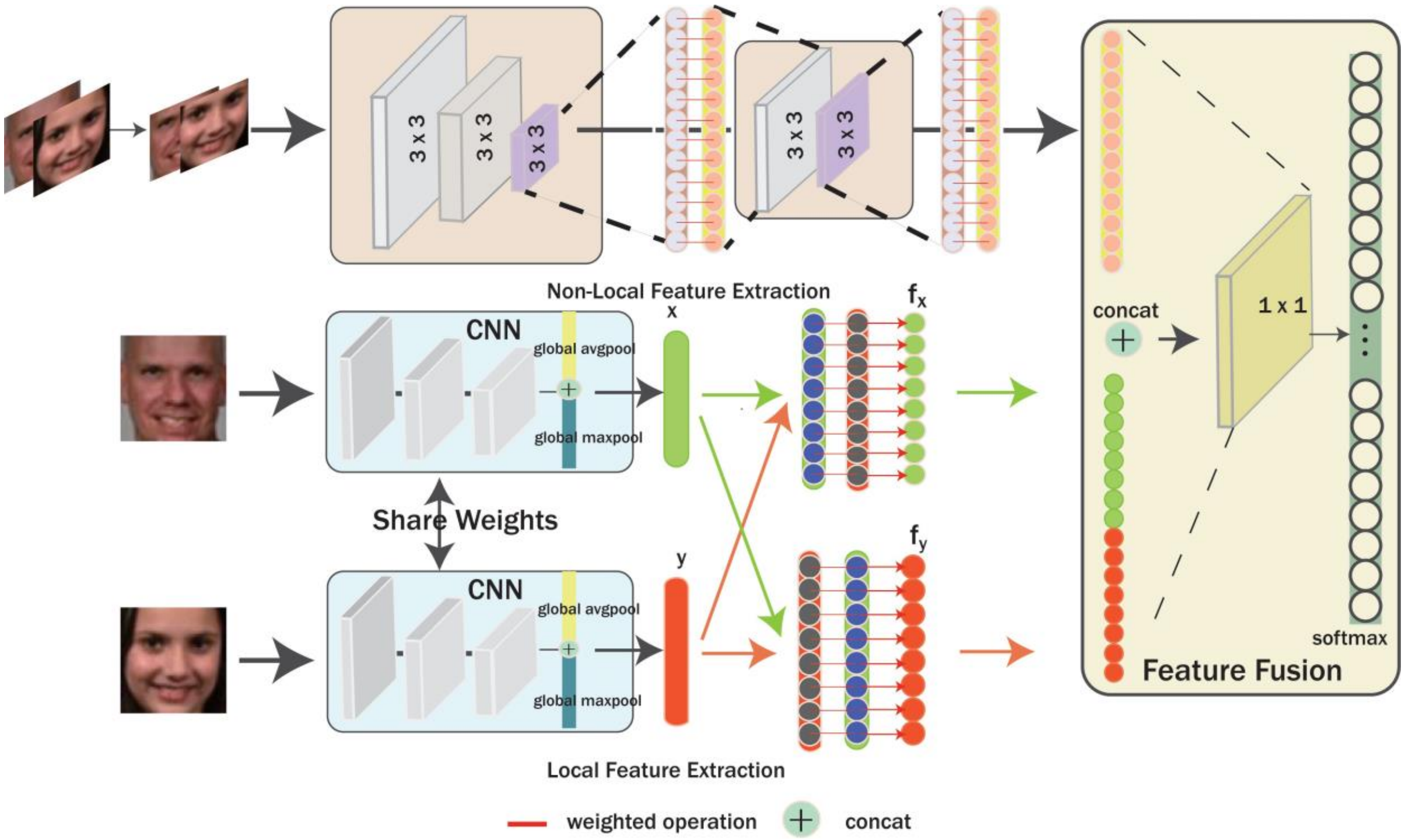

Fig. 2: The overview of our proposed framework. The non-local feature extraction module is conceived to learn the inter-dependent global information via leveraging the correlations across different images, while the local feature extraction module is constructed to produce the individual semantic-visual image representations according to their visual similarities.

of the face, such as texture and skin color, are extracted by different learning models.

Popular methods have been attempted to explore kinship, such as shallow learning algorithms [2], [4]–[6], [14]–[18], and deep-learning algorithms [7]–[9], [19]–[28]. The shallow algorithms can capture the most significant traits of each person, but they are hard to handle the nonlinear variations and rely on the pre-defined features. Notably, compared to shallow algorithms depending on experience, deep learning algorithms have powerful adaptability to dig up more potential and various characteristics. It shows an ability strong enough to capture the intrinsic structure hidden in the sample's knowledge explicitly. For example, an end-to-end shallow network [7] treated the kinship problems as a binary problem and unified the training process to deal with kin relations. Although the end-to-end learning algorithm does not require too much manual human intervention and has a simplified architecture, the current research on deep kinship verification is seldom trained in an end-to-end manner. For example, the method in [26] used CNNs for feature extraction, followed by the NRML [5] algorithm for metric learning. Moreover, a deep belief network [28] was designed to extract three non-overlapped regions in an unsupervised manner. Subsequently, the extracted features were fed into the support vector machine (SVM), the nearest neighbor (NN) [29] and the k-nearest neighbor (KNN) [29] for classification purposes. This case appears because the end-to-end learning networks require large amounts of labeled data, while the current datasets fail to meet the request for efficient and accurate verification.

Although encouraging results have been achieved, some limitations are still unsolved in the current research for kinship verification. As we know, deep learning can generate powerful feature representations for different vision-based tasks, but it still heavily relies on well-prepared training data. Moreover, the inherent problem of kinship verification is still under-explored. It is notable that intra-class variations in kinship verification are always higher than inter-class variations. However, it is still an open problem how could we construct effective metrics to maximize the intra-class compactness and inter-class separability of samples for discriminative kinship verification. As such, larger kinship datasets and more effective distance metrics are highly desirable to be developed to improve the performance of existing kinship verification methods.

### B. *Metric Learning*

Metric learning has been proposed to measure the similarity between the samples and achieved significant progress in computer vision, especially in few-shot learning. Typical methods can be roughly divided into linear transformation metric learning (such as linear discriminant analysis (LDA), marginal fisher analysis (MFA) [30], cosine similarity metric learning (CSML) [31], large margin nearest neighbor (LMNN) [32] and information theoretic metric learning (ITML) [33]) and nonlinear metric learning (such as Isometric Mapping (ISOMAP) [34], Locally Linear Embedding (LLE) [35] and Laplacian Eigenmap (LE) [36]). Generally, metric learning aims to minimize the distance between intra-class samples and maximize the distance between inter-class samples. How to measure the similarity between samples effectively is important to learn a good distance metric in different applications. With the advent of deep learning, considerable attentions have been paid to deep metric learning. Deep metric learning aims at applying a pre-determined analytical metric (such as Mahalanobis distance and Euclidean distance) into deep neural networks. For example, Chopra *et al.* [37] introduced a contrastive strategy in the Euclidean space to determine a large margin. FaceNet [38] further brings an advance and leverages a triplet loss in the Euclidean space to balance relative distance instead of absolute distance. An angular distance is employed in [39] to enforce a large margin between the positive and negative examples, so that it can increase the discriminative abilities of the learned features. Some other related works are center loss [40], lifted structured loss [41], multi-class n-pair loss [42], coupled clusters loss [43] and global loss [44].

Inspired by the above analysis, in this work, we develop a cross-generation feature interaction learning (CFIL) framework for robust kinship verification. Technically, the proposed method integrates feature learning and similarity metrics into one unified and end-to-end learning architecture, which could capture internal learning relations. Moreover, the information of interactions between two points is regarded as the interior auxiliary weights to mitigate the enormous divergence of intra-class variations, enhancing the discriminability of the learned features. Additionally, we also design a feature fusion strategy to obtain more complementary information by making full use of discriminative features contained in samples.

## III. Proposed Method

### A. Overview of the Proposed CFIL Method

To obtain more discriminative and comprehensive feature representations, in this paper, we propose a novel joint cross-generation feature interaction learning to compensate for the shortage of interior information extraction. Significantly, a collaborative weighting strategy is developed to explore comprehensive interactive relation details of images for robust kinship verification. As shown in Figure 2, this collaborative weighting strategy has two parts to find more internal correlations and details between parents-children facial images, *i.e.*, the non-local feature extraction module, and the local feature extraction module. The non-local part leverages the non-local weighted operation to guarantee the integrity of global information and learn the inter-dependent global information via leveraging the correlations across different images. Meanwhile, the local part introduces the local weighted operation to keep the individual semantic-visual image representations to guide and calibrate feature selection. When both the non-local and local parts adopt a collaborative weighting strategy in a mutually reinforced manner, this fusion strategy can more naturally express the semantics as much as possible and fully exploit consistency and complementary between images to improve the performance.

### B. Non-Local Feature Extraction

Kinship analyses are not only affected by features but also by internal correlations. To reflect the semantic relevance of global information and differentiate the discrete and independent features, we propose a non-local feature extraction to capture global relational information, in which the non-local weighted operation module can generate high-quality precise semantic representations by global semantic interaction.

Inspired by the non-local mean shown in [45], let $\mathbf{x} = \{x_i \mid i = 1, 2, \dots, n\} \in \mathbb{R}^n$ be a $n$-dimensional feature vector, and $i$ indexes the spatial location of the feature vector. Our non-local weighted operation can be regarded as an element-wise multiplication function shown below.

$$f_{x_i} = \sum_{m=1}^{n} w_{i,m} x_m, \tag{1}$$

where $w_{i,m}$ represents the non-local correlation between $x_i$ and $x_m$. The $f_{x_i}$ is the updated feature representation of $x_i$ and achieved by multiplying each location of the feature vector by its non-local correlation weights. Note that Eq. (1) can be updated in a matrix form:

$$\mathbf{f} = \mathbf{W}\mathbf{x}^T, \tag{2}$$

where $\mathbf{W} \in \mathbb{R}^{n\times n}$ denotes the accumulated correlation weight matrix, $\mathbf{f}$ is a semi-definite vector which has the same dimensions as $\mathbf{x}$.

Specifically, Eq. (1) and Eq. (2) satisfy the conditions $0 \le w_{i,m} \le 1$ and $\sum_{m=1}^{n} w_{i,m} = 1$. $m$ is the index that enumerates all spatial location of the feature vector and $1 \le m \le n$. To make a better express of interactive information, the non-local correlation between $x_i$ and $x_m$ can be defined as a similarity comparison extended by the Gaussian function shown in Eq. (3), where $\psi(\cdot)$ is the distance calculation. It should be mentioned that we tried the various combinations of distance, and $(x_i - x_m)^2 + (x_i^2 - x_m^2)$ can achieve promising performance.

$$w_{i,m} = \frac{1}{N_i} e^{\psi(x_i, x_m)}, \tag{3}$$

$$N_i = \sum_{m=1}^{n} e^{\psi(x_i, x_m)}. \tag{4}$$

where $N_i$ is a normalized factor for $x_i$, and it aims to promote structure information and control the effects between $x_i$ and outlying locations. By substituting (4) into (3), we have a total correlation function:

$$w_{i,m} = \frac{e^{\psi(x_i, x_m)}}{\sum_{m=1}^{n} e^{\psi(x_i, x_m)}}. \tag{5}$$

For all spatial locations $i = 1, 2, \dots, n$, the weight matrix can be extended as follows:

$$\mathbf{W} = \begin{bmatrix} w_{1,1}, w_{1,2}, \cdots, w_{1,n} \\ w_{2,1}, w_{2,2}, \cdots, w_{2,n} \\ \cdots, \cdots, \cdots, \cdots \\ w_{n,1}, w_{n,2}, \cdots, w_{n,n} \end{bmatrix} \tag{6}$$

The final $\mathbf{f}$ represents the updated feature representation which

---

**Algorithm 1** Calculation of Weights $\mathbf{W}$ in Eq. (6)

---

**Require:** The feature map with size $C \times H \times W$, initialize a Tensor $\mathbf{W}$ with size $n \times n$
**Ensure:** $n = C \times H \times W$
  Reshape the feature map $\mathbf{x}$ to size $1 \times n$
  **for** $i = 1 \ldots n$ **do**
    **for** $m = 1 \ldots n$ **do**
      $w_{i,m} = SoftMax((x_i - x_m)^2 + (x_i^2 - x_m^2))$, where $i, m \in n$
      $w_{i,m}$ is the weight between $x_i$ and $x_m$ in new feature space
    **end for**
  **end for**

---

projects feature vector from the original feature space to a new feature space shown in Eq. (7).

$$\mathbf{f} = \mathbf{W}\mathbf{x}^T, \\ = \begin{bmatrix} w_{1,1}, w_{1,2}, \cdots, w_{1,n} \\ w_{2,1}, w_{2,2}, \cdots, w_{2,n} \\ \cdots, \cdots, \cdots, \cdots \\ w_{n,1}, w_{n,2}, \cdots, w_{n,n} \end{bmatrix} \mathbf{x}^T. \tag{7}$$

In our work, we substitute $(x_i - x_m)^2 + (x_i^2 - x_m^2)$ into all formulas as follows:

$$\begin{cases} f_x = \frac{1}{N_i} \sum_{m=1}^{n} e^{((x_i - x_m)^2 + (x_i^2 - x_m^2))} x_m \\ \mathbf{W} = \{w_{i,1}, w_{i,2}, \cdots, w_{i,n}\} \\ N_i = \sum_{m=1}^{n} e^{(x_i - x_m)^2 + (x_i^2 - x_m^2)} \end{cases} \tag{8}$$

The main structure of the non-local feature extraction is shown in Figure 2. The corresponding learning algorithm is summarized in Algorithm 1. Figure 3 illustrates the detailed weighted operation of non-local feature extraction. The training is performed in a modular fashion where the first three convolutional layers learn a high-level feature map. The feature map is reshaped to $1 \times CHW$. Subsequently, the non-local weighted operation learns the pairwise cross-generation interaction information and projects the vector from the original feature space into a new feature space. And then, the new feature map is reshaped to size $C \times H \times W$, which is considered as the input of the next layer. The final convolutional layer aggregates the cumulative two types of kinship representations to compensate for the lack of information. The final classifier is a combination of two fully-connected layers and a softmax layer.

### C. Local Feature Extraction

To make full use of the semantic consistency and similarities between image pairs, we introduce a local feature extraction module achieved by a local weighted operation to capture and preserve the details of the single image.

As shown in Figure 2, we employ two popular CNN models, i.e., VGG-19 [46] and Resnet-50 [47], as the backbones for local feature extraction. Each model is pre-trained on ImageNet [48]. All backbones have been frozen all weights and bias by the back-propagation procedure of training data. Comparing to ImageNet, our datasets are too small to retrain the model. Therefore, local feature extraction leverages the weights learning from ImageNet to make the network have more abilities to fit our datasets. The model removes the last fully-connected layer and classifier by replacing them with two new parallel global pooling layers (GlobalAvgpooling with kernel size $7 \times 7$ and GlobalMaxpooling with kernel size $7 \times 7$, both have the same size as last feature map) instead of flattening directly. Global pooling is easier to switch between feature maps and final classification. Unlike fully-connected layers, since global pooling does not require a large number of parameters for training and tuning, such a strategy will make the model more robust, and the anti-overfitting will be better. Moreover, the new layer can learn the global features we need automatically.

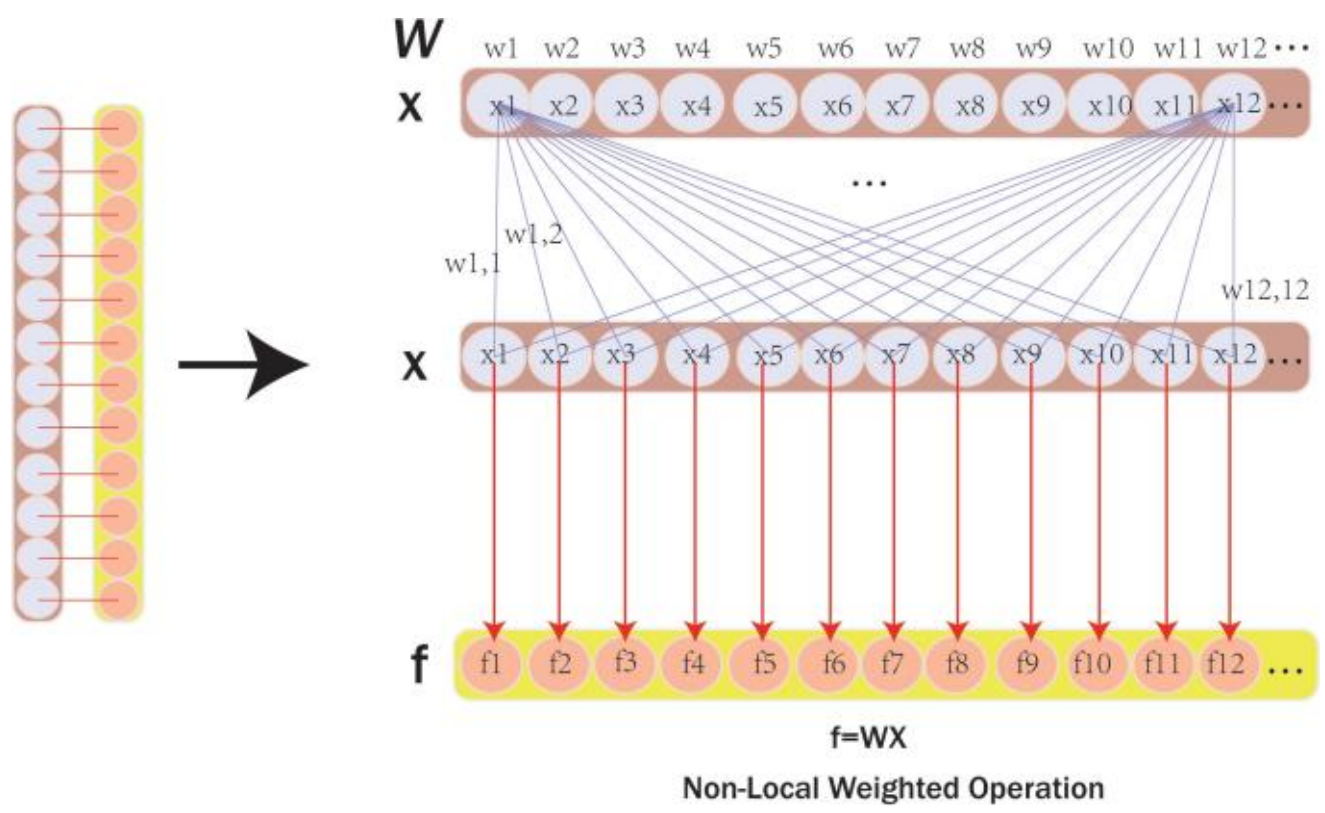

Fig. 3: The main module of the non-Local weighted operation.

The original spatial resolution is $64 \times 64$, but in our local feature extraction, we resize it to size $224 \times 224$. We consider a pair image of parent-child face sets as the inputs for this network. After the backbone, we can get a feature map with size (2048,7,7) of Resnet-50 or (512,7,7) of VGG-19 for each image.

The detailed structure of the local feature extraction is similar to the non-local one, except for the main objective functions at each image. Moreover, given two different feature vectors $\mathbf{x} = \{x_i \mid i = 1, 2, \ldots, n\} \in \mathbb{R}^n$, $\mathbf{y} = \{y_j \mid j = 1, 2, \ldots\} \in \mathbb{R}^n$, where $i$ and $j$ index the spatial location of $\mathbf{x}$ and $\mathbf{y}$, respectively. For each location, our local weighted operation can be denoted as:

$$\begin{cases} f_{x_i} = \sum_{k=1}^{n} w_{i,j} x_k \\ f_{y_j} = \sum_{k=1}^{n} w_{j,i} y_k \end{cases}, \tag{9}$$

where $w_{i,j}$ represents the local correlation between $x_i$ and $y_j$, $w_{j,i}$ represents the local correlation between $y_j$ and $x_i$. The $f_{x_i}$ and $f_{y_j}$ are the updated feature representations of $x_i$ and $y_j$, respectively. $k$ is the index that enumerates all spatial location of feature vector. The local correlation $w$ can be defined as:

$$\begin{cases} w_{i,j} = \frac{1}{N_{x_i}} e^{\psi(x_i, y_j)} \\ w_{j,i} = \frac{1}{N_{y_j}} e^{\psi(y_j, x_i)} \end{cases}, \tag{10}$$

$$\begin{cases} N_{x_i} = \sum_{j=1}^{n} e^{\psi(x_i, y_j)} \\ N_{y_j} = \sum_{i=1}^{n} e^{\psi(y_j, x_i)} \end{cases}. \tag{11}$$

where all $w$ satisfy the same conditions as non-local operation, $N_{x_i}$ and $N_{y_j}$ are normalized factors for $x_i$ and $y_j$, respectively. $\psi(\cdot)$ is the distance calculation.

For all spatial locations in $\mathbf{x}$ and $\mathbf{y}$, the local correlation weight matrices can be extended as follows:

$$\mathbf{W_x} = \begin{bmatrix} w_{1,1}, w_{1,2}, \cdots, w_{1,n} \\ w_{2,1}, w_{2,2}, \cdots, w_{2,n} \\ \cdots, \cdots, \cdots, \cdots \\ w_{n,1}, w_{n,2}, \cdots, w_{n,n} \end{bmatrix}, \tag{12}$$

$$\mathbf{W_y} = \begin{bmatrix} w_{1,1}, w_{1,2}, \cdots, w_{1,n} \\ w_{2,1}, w_{2,2}, \cdots, w_{2,n} \\ \cdots, \cdots, \cdots, \cdots \\ w_{n,1}, w_{n,2}, \cdots, w_{n,n} \end{bmatrix}. \tag{13}$$

where $\mathbf{W_x} \in \mathbb{R}^{n\times n}$ and $\mathbf{W_y} \in \mathbb{R}^{n\times n}$ denote the accumulated local correlation weight matrix of $\mathbf{x}$ and $\mathbf{y}$, respectively. Meanwhile, Eq. (9) can be updated in a matrix form:

$$\begin{cases} \mathbf{f_x} = \mathbf{W_x}\mathbf{x}^T \\ \mathbf{f_y} = \mathbf{W_y}\mathbf{y}^T \end{cases} \tag{14}$$

where $\mathbf{f_x}$ and $\mathbf{f_y}$ are semi-definite vectors which have the same dimensions as $\mathbf{x}$ and $\mathbf{y}$, respectively.

In our paper, we substitute the distance function $(x_i - y_j)^2 + (x_i^2 - y_j^2)$(feature vector $\mathbf{x}$) and $(y_j - x_i)^2 + (y_j^2 - x_i^2)$(feature vector $\mathbf{y}$) into all representations as follows:

$$\begin{cases} f_{x_i} = \frac{1}{N_{x_i}} \sum_{k=1}^{n} e^{((x_i - y_j)^2 + (x_i^2 - y_j^2))} x_k \\ f_{y_j} = \frac{1}{N_{y_j}} \sum_{k=1}^{n} e^{((y_j - x_i)^2 + (y_j^2 - x_i^2))} y_k \\ \mathbf{W_x} = \{ w_{i,1}, w_{i,2}, \cdots, w_{i,n} \} \\ \mathbf{W_y} = \{ w_{j,1}, w_{j,2}, \cdots, w_{j,n} \} \\ N_{x_i} = \sum_{j=1}^{n} e^{((x_i - y_j)^2 + (x_i^2 - y_j^2))} \\ N_{y_j} = \sum_{i=1}^{n} e^{((y_j - x_i)^2 + (y_j^2 - x_i^2))} \end{cases}. \tag{15}$$

The explicit learning steps of the local weighted operation are shown in Algorithm 2. Figure 4 illustrates the detailed weighted operation of local feature extraction. A pair of images with $224 \times 224$ spatial resolution goes through convolutional layers for extracting a high-level feature map. And then, we reshape the feature map to size $1 \times CHW$ so that the local weighted operation can learn another type of cross-generation interaction information. After the local weighted operation, we concatenate two semantic vectors and non-local features as the input of $1*1$ convolutional layer to reduce the parameters, computation, and capacity of the framework for further recognition.

**Algorithm 2** Calculation of Weights $\mathbf{W_x}, \mathbf{W_y}$ in $(12, 13)$

**Require:** The two feature maps with size $C \times H \times W$, initialize two Tensors $\mathbf{W_x}$ ,$\mathbf{W_y}$ with size $n \times n$
**Ensure:** $n = C \times H \times W$
Reshape two feature maps to $\mathbf{x} = \{x_i \mid i=1,2,\ldots, n\}$, $\mathbf{y} = \{y_j \mid j=1,2,\ldots,n\}$
**for** $i = 1 \cdots n$ **do**
  **for** $j = 1 \cdots n$ **do**
    $w_{i,j} = SoftMax((x_i - y_j)^2 + (x_i^2 - y_j^2))$, Where $i, j = 1, 2, \ldots, n$
    $w_{i,j}$ is the weight between $x_i$ and $y_j$ in new feature space
  **end for**
**end for**
**for** $j = 1 \cdots n$ **do**
  **for** $i = 1 \cdots n$ **do**
    $w_{j,i} = SoftMax((y_j - x_i)^2 + (y_j^2 - x_i^2))$, Where $i, j = 1, 2, \ldots, n$
    $w_{j,i}$ is the weight between $y_j$ and $x_i$ in new feature space
  **end for**
**end for**

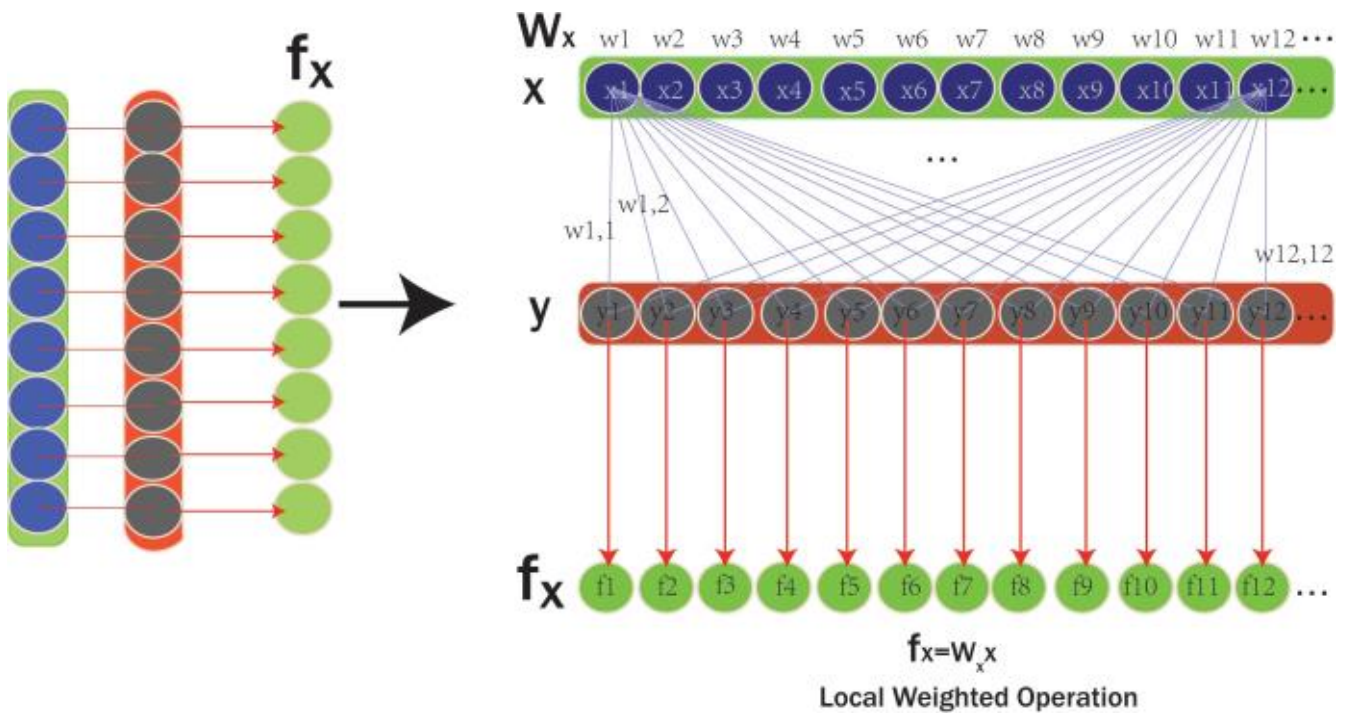

Fig. 4: The main module of the Local weighted operation.

## D. Loss Function

We train our model by using softmax cross-entropy loss function to supervise this task. Given a kinship dataset, we can denote $f_i$ as the final feature of $ith$ sample. $r_i \in [0, 1]$ is the ground truth.

The loss function can be written as:

$$L^i_{main} = -\log P_i, P_i = \frac{e^{h(f_i)}}{\sum_{r_i=0}^{1} e^{h_{r_i}(f_i)}} \tag{16}$$

where $P_i$ is the final predicted probability distribution over sources of the $i$th sample. $h(.) = \theta^T f_i + b_i$, where $\theta$ is a set of the weights that contain different classes of the $i$th sample. The derivative of $L^i_{main}$ with respect to $h$ is defined as follows:

$$\begin{aligned} \frac{\partial L^i_{main}}{\partial h(f_i)} &= \frac{\partial L^i_{main}}{\partial P_i} \frac{\partial P_i}{\partial h(f_i)} \\ &= -\frac{1}{h(f_i)} \frac{\partial P_i}{\partial h(f_i)}. \end{aligned} \tag{17}$$

For the softmax loss, the derivative of $P_i$ with respect to $h$ is formulated as follows:

$$\frac{\partial P_i}{\partial h(f_i)} = \frac{\partial P_i(l \mid f_i)}{\partial h_l(f_i)} = \begin{cases} -P_i(l \mid f_i)P_i(l \mid f_i), & l \neq r_i \\ P_i(l \mid f_i)(1 - P_i(l \mid f_i)), & l = r_i \end{cases}, \tag{18}$$

where $P(l \mid f_i)$ is the final predicted probability of the $i$th sample when the prediction is $l$.

By substituting (18) into (17), we can get

$$\frac{\partial L_i}{\partial h(x_i)} = \begin{cases} P_i(l \mid x_i), & l \neq r_i \\ P_i(l \mid x_i) - 1, & l = r_i. \end{cases} \tag{19}$$

Finally, we use a total loss function to supervise the proposed generic framework, and we have

$$\begin{aligned} L &= \frac{1}{\mathbf{N}} \sum_{i=1}^{\mathbf{N}} L^i_{main} \\ &= \frac{1}{\mathbf{N}} \sum_{i=1}^{\mathbf{N}} -\log P_i \\ &= \frac{1}{\mathbf{N}} \sum_{i=1}^{\mathbf{N}} -\log \frac{e^{h(f_i)}}{\sum_{r_i=0}^{1} e^{h_{r_i}(f_i)}}, \end{aligned} \tag{20}$$

where $\mathbf{N}$ is the mini-batch size of the training samples.

## IV. Experiments and Results

In this section, to validate the effectiveness of the proposed method, we conduct extensive experiments on four widely-used kinship datasets by comparing some state-of-the-art kinship verification algorithms.

### A. Datasets

There are many facial datasets for kinship verification. We use four challenging kinship datsets to evaluate the performance of different algorithms, including **Large-Scale Families in the Wild (FIW)** dataset [49], [50] and Small-Scale Kin datasets: **KinFaceW-I** [5], **KinFaceW-II** [5], **UBKinFace** [51] as training and testing the network as shown in Table I. The detailed description of each dataset is illustrated as follows.

**FIW** [49] is the largest dataset for kinship verification. This dataset includes approximately 656K face pairs of 10676 people collected from 1000 different families. There are 11 different kin relations.

**KinFaceW-I** [5] is a dataset of unconstrained face images that consists of 1066 samples collected from 533 people. There are 4 different kin relations. This dataset is selected from different people, and it is contaminated by noise. There are four representative types of kin relations, i.e., Father-Son (F-S), Father-Daughter (F-D), Mother-Son (M-S) and Mother-Daughter (M-D). There are 156, 134, 116, and 127 pairs in KinFaceW-I.

**KinFaceW-II** [5] is also a face image dataset pictured under unconstrained conditions. This dataset has 2000 samples collected from 1000 people. There are 4 different kin relations. The difference between KinFaceW-II [5] and KinFaceW-I [5] is that each parent-child pair of KinFaceW-II is selected from the same photos. Similarly, there are four representative types of kin relations: Father-Son (F-S), Father-Daughter (F-D), Mother-Son (M-S), and Mother-Daughter (M-D), respectively. There are 250 pairs of each kin relation in KinFaceW-II.

TABLE I: Family-based characteristics in experiments.

| Dataset | No. family | No. people | No. samples | Kin relations | Multiple images |
|---|---|---|---|---|---|
| FIW [49] | 1,000 | 10,676 | 656,000 | 11 | Yes |
| KinFaceW-I [5] | – | 533 | 1,066 | 4 | No |
| KinFaceW-II [5] | – | 1,000 | 2,000 | 4 | No |
| UB KinFace [51] | – | 400 | 600 | 4 | No |

**UBKinFace** [51] consists of young children, their young parents and old parents. This dataset contains 200 triplets and includes more than 1,000 images. Typically, most of them are real collections from public figures (celebrities and politicians) from the Internet and have been widely used to verify kinship.

### B. Baselines

Some representative methods are employed to testify the performance on different datasets under the same experimental configurations, such that the experimental results are reliable and convincing. We simply list the description on each algorithm as follows:

*1) Shallow learning-based models:*

- **ASML** [16]: This method employs an adversarial metric learning to build a similarity metric.
- **LDA, MFA, WGEML** [30]: Some conventaional feature learning models such as LDA (Linear Discriminant Analysis), MFA (Marginal Fisher Analysis) and WGEML (Weighted Graph Embedding Based Metric Learning) are used to get multiple similarity metrics.
- **DMML** [6]: This method uses multi-features derived from different descriptors to learn multiple distance metrics.
- **NRML** [5]: The method uses NRML (Neighborhood Repulsed Metric Learning) to determine a distance metric.
- **L$M^3$L** [15]: This method uses multiple feature descriptors to extract various features for each face image.
- **DDMML** [17]: This method proposes a discriminative deep multi-metric learning method to make better use of different feature descriptors and maximize the correlation of different features for each sample.
- **MKSM** [18]: This method uses a multiple kernel similarity metric (MKSM) to combine multiple basic similarities for the feature fusion.
- **KINMIX** [52]: This method verifying kin relations by using a KinMix method to generate positive samples for data augmentation.

*2) Deep learning-based models:*

- **CNN-Basic, CNN-Points** [7]: Both models uses the deep CNN model to extract features and classify the kin relation.
- **SMCNN** [8]: This method uses the similarity metric based CNNs to verify kin relation.

- **DTL** [27]: This method uses a transfer learning strategy and triangular similarity metric to train model, and leverages both face and the mirror face to increase robustness and accuracy.
- **CFT** [26]: This method combines transfer learning-based CNN and metric learning (NRML or other metrics) to get the final features.
- **DKV** [9]: This method uses the LBP features as the first input of an auto-encoder network, and then uses a metric learning for prediction.
- **fcDBN** [28]: This method uses deep belief networks to extract hierarchical feature representation by three different regions of faces. This model can achieve state-of-the-art performances on different datasets.
- **AdvKin,E-AdvKin** [53]: This method uses a adversarial convolutional network with residual connections for facial kinship verification.
- **GKR** [54]: This method employs a graph-based kinship reasoning (GKR) network for kinship verification.
- **SMNAE** [55]: This method utilizes the learned spatio-temporal representation in the video for verifying kinship in a pair of videos.
- **ResNet+SDMLoss** [56]: This method mitigates both age and identity divergences between cross-generation for effective kinship verification.
- **SphereFace** [50]: This method uses the angular softmax (A-Softmax) loss to learn angularly discriminative features for face verification.
- **VGG+DML** [57]: This method proposes a de-noising auto-encoder based robust metric learning (DML) framework to learn discriminative features.

***Remark 1***: It should be noted that **SMNAE** [55] is regarding kinship verification on videos.

### C. Experimental Settings

For evaluating the performance of the proposed method, all experiments train on four widely-used datasets, i.e., the FIW [49], [50], KinFaceW-I [5], KinFaceW-I [5] and UBKinFace [51] datasets. For performance validation, we use MTCNN [58] to detect and align the face region and only conduct central cropping into 73 by 73, so that the final spatial resize of the inputs is 64 by 64 after data augmentation. All face images are aligned and cropped into $64 \times 64$ and $224 \times 224$ two types of face for CFIL. Table II shows the details of our structure. Especially for the non-local feature extraction module, it has only one input terminal but can input two images, and it does not need to perform the necessary feature maps transformation. Specifically, we concatenate the image pair in a parallel manner and take them as one input with size $64 * 64 \times 6$ to handle two original parallel data. Hence, the original channels will be changed from 3 to 6. Notably, we perform five-fold cross-validation on all datasets. Each fold contains positive samples (with kinship relation) and negative samples (without kinship relation). We keep the images in all relationships to be roughly equal in all folds.

Following the existing works [5], [9], [28], in our experiments, each parent matches their children randomly for negative samples, who are not the parents' real children. Moreover, each image of the parent-offspring pair only uses once in the negative samples. All datasets are divided into two groups: face images with kinship relation are used as positive samples, and those without kinship relation are regarded as negative samples. All experiments follow the $"80\% - 20\%"$ protocol: we randomly select 80% sample images of this database for training, and the remained 20% images are used for testing. We train our kinship network by utilizing Adam optimizer with a batch size 64. The initial learning rate is set to 0.001, and we decrease it to 0.0005 after the second epoch. The momentum is set to 0.9.

TABLE II: Backbone Structure

| Non-local | Local(ResNet-50) | Local(VGG-19) |
|---|---|---|
| conv, $3 \times 3, 16$, **stride**1<br>max pool, $2 \times 2$, **stride**1<br>$(32 \times 32)$ | **conv**, $7 \times 7, 64$, **stride**2<br>max pool, $3 \times 3$, **stride**2<br>$(112 \times 112)$ | conv, $3 \times 3, 64$<br>conv, $3 \times 3, 64$ |
| conv, $3 \times 3, 64$, **stride**1<br>max pool, $2 \times 2$, **stride**1<br>$(16 \times 16)$ | $\begin{bmatrix} conv & 1 \times 1, & 64 \\ conv & 3 \times 3, & 64 \\ conv & 1 \times 1, & 256 \end{bmatrix} \times 3$<br>$(56 \times 56)$ | conv, $3 \times 3, 128$<br>conv, $3 \times 3, 128$ |
| conv, $3 \times 3, 128$, **stride**1<br>max pool, $2 \times 2$, **stride**1<br>**Non-local feature extraction**$(8 \times 8)$ | $\begin{bmatrix} conv & 1 \times 1, & 128 \\ conv & 3 \times 3, & 128 \\ conv & 1 \times 1, & 512 \end{bmatrix} \times 4$<br>$(28 \times 28)$ | conv, $3 \times 3, 256$<br>conv, $3 \times 3, 256$<br>conv, $3 \times 3, 256$<br>conv, $3 \times 3, 256$ |
| conv, $3 \times 3, 256$, **stride**1<br>max pool, $2 \times 2$, **stride**1<br>$(4 \times 4)$ | $\begin{bmatrix} conv & 1 \times 1, & 256 \\ conv & 3 \times 3, & 256 \\ conv & 1 \times 1, & 1024 \end{bmatrix} \times 6$<br>$(14 \times 14)$ | conv, $3 \times 3, 512$<br>conv, $3 \times 3, 512$<br>conv, $3 \times 3, 512$<br>conv, $3 \times 3, 512$ |
| conv, $3 \times 3, 512$, **stride**1<br>max pool, $2 \times 2$, **stride**1<br>**Non-local feature extraction**$(2 \times 2)$ | $\begin{bmatrix} conv & 1 \times 1, & 512 \\ conv & 3 \times 3, & 512 \\ conv & 1 \times 1, & 2048 \end{bmatrix} \times 3$<br>$(7 \times 7)$ | conv, $3 \times 3, 512$<br>conv, $3 \times 3, 512$<br>conv, $3 \times 3, 512$<br>conv, $3 \times 3, 512$<br>$(7 \times 7)$ |
| | global average pool, $7 \times 7$<br>+<br>global max pool, $7 \times 7$<br>$(1 \times 1)$ | global average pool, $7 \times 7$<br>+<br>global max pool, $7 \times 7$<br>$(1 \times 1)$ |
| 512*2*2 | 2048*1*1+2048*1*1 | (512*1*1+512*1*1) |

We employ two popular CNN models, i.e., VGG-19 [46] and Resnet-50 [47], as the backbones for local feature extraction. Each model is pre-trained on ImageNet [48]. All backbones have been frozen all weights and bias by the back-propagation procedure of training data. After the backbone, we can get a feature map with size (2048,7,7) of Resnet-50 or (512,7,7) of VGG-19 for each image. It is necessary to mention that our collaborative weighting strategy does not change the original feature size and is unchanged for implementation. Specifically, according to Table II backbone structure and algorithm 1 & 2, during the training and testing, the $n$ of correlation matrices $\mathbf{W}$ are set as $8192(128 * 8 \times 8)$ and $1024(512 * 2 \times 2)$ for non-local feature extraction, $4096$ (ResNet-50) and $1024$ (VGG-19) for local feature extraction. The final prediction layer channel is set to 2. We train our network on the first three folds of the FIW [49] dataset, and then use the fourth fold to tune the parameters. The evaluation results are directly cited from the corresponding original papers for fair comparisons.

### D. Evaluation Metrics

We compare our method with the state-of-the-art algorithms on the Mean Verification Accuracy (MVA) score. We use the weighted accuracy (WA) and MVA to represent our main overall accuracy. Moreover, we use Mean Verification Accuracy (MVA) as our average accuracy. To make an intuitive comparison of our method and other algorithms, we exploit the ROC (the receiver operating characteristic) curve on these four datasets. ROC curve has been widely used in kinship verification. The ROC metric is plotted with the x-axis (FPR) and y-axis (TPR). FPR is False Positive Rate, and TPR is a True Positive Rate. The definitions are given as follows.

$$TPR = \frac{TP}{TP + FN} \tag{21}$$

$$FPR = \frac{FP}{TN + FP} \tag{22}$$

$$ACC = \frac{TP + TN}{P + N} = \frac{TP + TN}{TP + TN + FP + FN} * 100\%, \tag{23}$$

$$WA = \frac{1}{2}\left(\frac{TP}{P} + \frac{TN}{N}\right) * 100\%. \tag{24}$$

where TN, TP, FP, and FN denote the true negative, true positive, false positive, and false negative, respectively. For example, in a binary classification task, the results are labeled either as positive (p) or negative(n). If the prediction is p and the real value is p, then it is true positive (TP). However, if the real value is n, then it is a false positive (FP). TN (true negative) means both the prediction and the real value are n, and FN (false negative) means the prediction is n while the real value is p. The WA is widely used in dealing with imbalanced data. The ROC curve indicates that the greater the area under the curve, the accuracy is higher. Since the source codes of most compared algorithms are unavailable and they are hard to be reproduced, we only draw our methods. Notably, some methods do not have the ROC curve, and we directly compare their Mean Verification Accuracy scores with our method. Moreover, for overall accuracy, we use WA as our primary evaluation metric, and MVA as our secondary evaluation metric.

### E. Experimental Results and Analysis

Our experiments are tested on four public datasets.

*1) Proposed Method:* Table III, IV, Figure 5, Figure 6 and Figure 7 show our results on four datasets. It is clear that the performance of our CFIL is better than all the compare algorithms in most cases, especially CFIL-Resnet50 on all datasets. Our model can catch more details and better feature representations. Figure 5 shows the ROC curves of the proposed method. Figure 7 shows the verification rate of CFIL versus different number of iteration on different kinship datasets. From the experimental results, we can see that our proposed method appears a local optimal peak in the first several iterations.

*2) Comparison With Shallow Learning-Based Models:* As shown in Table III, we compare our algorithms with shallow learning-based methods. The proposed model consistently outperforms state-of-the-art shallow-learning-based methods, i.e. ASML [16], MLDA [30], MMFA [30], WGEML [30], DMML [6], NRML [5], L$M$ $^3$L [15], DDMML [17], MKSM [18], and KINMIX [52]. The table shows that the highest accuracy is obtained on KinFaceW-II subset Mother-daughter(MD). We have at least a 12% gain on the KinfaceW-I dataset. Besides, we have at least an 8% gain on the KinfaceW-II dataset. We have better performance on the UB Kin dataset, and at least a 17% gain on overall accuracy is achieved to verify the effectiveness of the proposed method.

The effectiveness of CFIL has been demonstrated. Table III shows that CFIL can get better performance than the shallow learning-based models. The possible reason is that these methods are based on handcrafted features, which are relatively complex to verify. The selection of features depends mostly on experience, and its tune requires much time. Moreover, the features extracted from handcrafted methods are not efficient in complicated classification. Hence, it is necessary to select a classifier with superior performance. In particular, these methods need manually construct the distance function by selecting the appropriate feature for a specific task. However, this approach requires a large manual input and can be very inrobust to changes in data so that these features lack sufficient representation abilities for big data, leading to inferior performance for complicated computer vision tasks. By contrast, deep learning-based methods show powerful superiority in image feature extraction and classification compared to the conventional methods. These methods extract features leveraging pre-trained models that have powerful representation capabilities to capture more complementary information.

*3) Comparison With Deep Learning-Based Models:* As shown in Table III, Table IV, we compare our algorithms with state-of-the-art deep learning-based models. Compared to these models, our CFIL-Resnet50 Network shows better performance than state-of-the-art results. That is because Resnet-50 is deeper than VGG-19 and can be used to extract more complex features. It has better utilized residual learning of face images to solve the degradation problem and provide more discriminative information for kinship relation discovery. Our method achieves an improvement of 0.3% on the KinFaceW-I dataset, while a performance enhancement of 1% on the KinFaceW-II dataset. The results obtained on the KinFaceW-II dataset are generally higher than the KinFaceW-I dataset that is because the images of KinFaceW-II are selected from the same photos which have the same feature distribution space, while the images from the KinFaceW-I are disturbed by the illumination and aging variations. We improve the performance on the UB Kin dataset for a total 0.2% gain on the overall accuracy. Moreover, we enhance the performance on the FIW dataset for a 13% gain on overall accuracy.

The superior performances of our method may benefit from the following reasons. First, compared to the end-to-end method(such as CNN-Basic [7]), our method captures more comprehensive information by interaction learning. It increases the discriminability of the learned to reduce dif-overall : left is mean verification accuracy of all subsets, red is weighted accuracy.

TABLE III: Mean Verification Accuracies (%) of Previous Models for 5-Fold Experiment on Different Datasets.

| Learning Strategy | Methods | KinFaceW-I | | | | | KinFaceW-II | | | | | UB KinFace | | | | |
|---|---|---|---|---|---|---|---|---|---|---|---|---|---|---|---|---|
| | | F-S | F-D | M-S | M-D | overall | F-S | F-D | M-S | M-D | overall | F-S | F-D | M-S | M-D | overall |
| Shallow Learning | ASML [16] | 82.7 | 76.1 | 78.0 | 81.6 | 79.6 | 85.8 | 78 | 81.8 | 78.6 | 81.1 | – | – | – | – | – |
| | MLDA [30] | 76.6 | 71.2 | 77.7 | 76.4 | 75.5 | 86.6 | 74.4 | 81.0 | 78.8 | 80.2 | – | – | – | – | – |
| | MMFA [30] | 77.9 | 72.0 | 77.2 | 75.2 | 75.6 | 85.6 | 73.2 | 80.4 | 77.2 | 79.1 | – | – | – | – | – |
| | NRML [5] | 76.3 | 69.8 | 77.2 | 75.9 | 74.8 | 82.6 | 68.6 | 76.6 | 73.0 | 75.2 | – | – | – | – | – |
| | WGEML [30] | 78.5 | 73.9 | 80.6 | 81.9 | 78.7 | 88.6 | 77.4 | 83.4 | 81.6 | 82.8 | – | – | – | – | – |
| | DMML [6] | 74.5 | 69.5 | 69.5 | 75.5 | 72.3 | 78.5 | 76.5 | 78.5 | 79.5 | 78.3 | – | – | – | – | 72.25 |
| | DDMML [17] | 86.4 | 79.1 | 81.4 | 87.0 | 83.5 | 87.4 | 83.8 | 83.2 | 83.0 | 84.3 | – | – | – | – | – |
| | $LM^3L$ [15] | – | – | – | – | – | 82.4 | 74.2 | 79.6 | 78.7 | 78.7 | – | – | – | – | – |
| | MKSM [18] | 83.65 | 81.35 | 79.69 | 81.16 | 81.46 | 83.80 | 81.20 | 82.40 | 82.40 | 82.45 | – | – | – | – | – |
| | KINMIX [52] | 75.6 | 76.5 | 78.5 | 83.5 | 78.5 | 89.6 | 87.2 | 91.2 | 90.6 | 89.7 | – | – | – | – | – |
| Deep Learning | CNN-Basic [7] | 75.7 | 70.8 | 73.4 | 79.4 | 74.83 | 84.9 | 79.6 | 88.3 | 88.5 | 85.33 | – | – | – | – | – |
| | CNN-Points [7] | 76.1 | 71.8 | 78.0 | 84.1 | 77.5 | 89.4 | 81.9 | 89.9 | 92.4 | 88.4 | – | – | – | – | – |
| | SMCNN [8] | 75.0 | 75.0 | 68.7 | 72.2 | 72.73 | 75.0 | 79.0 | 78.0 | 85.0 | 79.25 | – | – | – | – | – |
| | DTL [27] | 86.45 | 89.62 | 89.57 | 88.8 | 88.61 | 83.4 | 89.4 | 87.0 | 87.0 | 86.7 | – | – | – | – | – |
| | CFT [26] | 78.8 | 71.7 | 77.2 | 81.9 | 77.4 | – | – | – | – | 65.5 | 77.4 | 76.6 | 79.0 | 83.8 | 79.3 |
| | DKV [9] | 71.8 | 62.7 | 66.4 | 66.6 | 66.9 | 73.4 | 68.2 | 71.0 | 72.8 | 71.3 | – | – | – | – | 72.25 |
| | fcDBN [a][28] | 98.1 | 96.3 | 90.5 | 98.4 | 95.825 | 96.8 | 94.0 | 97.2 | 96.8 | 96.2 | – | – | – | – | 91.75 |
| | AdvKin [53] | 75.7 | 78.3 | 77.6 | 83.1 | 78.7 | 88.4 | 85.8 | 88.0 | 89.8 | 88.0 | – | – | – | – | 81.4 |
| | E-AdvKin [53] | 76.6 | 77.3 | 78.4 | 86.2 | 79.6 | 91.6 | 85.2 | 90.2 | 92.4 | 89.9 | – | – | – | – | 80.4 |
| | GKR [54] | 79.5 | 73.2 | 78.0 | 86.2 | 79.2 | 90.8 | 86.0 | 91.2 | 94.4 | 90.6 | – | – | – | – | – |
| | SMNAE [b][55] | – | – | – | – | 96.9 | – | – | – | – | 97.1 | – | – | – | – | 95.3 |
| Deep Learning | CFIL-VGG19 | 92.2 | 93.2 | 93.5 | 93.0 | 93.0\93.2 | 92.8 | 93.0 | 94.3 | 93.8 | 93.5\93.5 | 88.4 | 89.3 | 90.7 | 91.2 | 89.9\88.9 |
| | CFIL-ResNet50 | 97.5 | 95.6 | 96.8 | 94.3 | 96.1\97.2 | 96.3 | 97.6 | 96.9 | 98.0 | 97.2\97.2 | 90.2 | 91.5 | 90.4 | 92 | 91\90.7 |

[a] the-state-of-the-art.

[b] It should be noted that **SMNAE** [55] is regarding kinship verification on videos.

TABLE IV: Mean Verification Accuracies (%) for 5-Fold Experiment on FIW with No Family Overlap Between Folds.

| Learning Strategy | Methods | siblings | | | parent-child | | | | grandparent-grandchild | | | | |
|---|---|---|---|---|---|---|---|---|---|---|---|---|---|
| | | B-B | S-S | SIBS | F-D | F-S | M-D | M-S | GF-GD | GF-GS | GM-GD | GM-GS | overall |
| Deep Learning | SphereFace [50] | 71.94 | 77.30 | 70.23 | 69.25 | 68.50 | 71.81 | 69.49 | 66.07 | 66.36 | 64.58 | 65.40 | 69.18 |
| | VGG+DML [57] | – | – | 75.27 | 68.08 | 71.03 | 70.36 | 70.76 | 64.90 | 64.81 | 67.37 | 66.50 | 68.79 |
| | ResNet+SDMLoss [56] | – | – | – | 69.02 | 68.60 | 72.28 | 69.59 | 65.89 | 65.12 | 66.41 | 64.90 | 69.47 |
| Deep Learning | CFIL-VGG19 | 86.6 | 81.4 | 79.8 | 87.5 | 91.6 | 89.4 | 88.8 | 77.2 | 75.4 | 76.9 | 72.4 | 82.5\83.8 |
| | CFIL-ResNet50 | 89.5 | 83.3 | 81.9 | 89.3 | 91.8 | 90.2 | 89.7 | 78.6 | 75.3 | 69.8 | 73.3 | 83\88.9 |

overall : left is mean verification accuracy of all subsets, red is weighted accuracy.

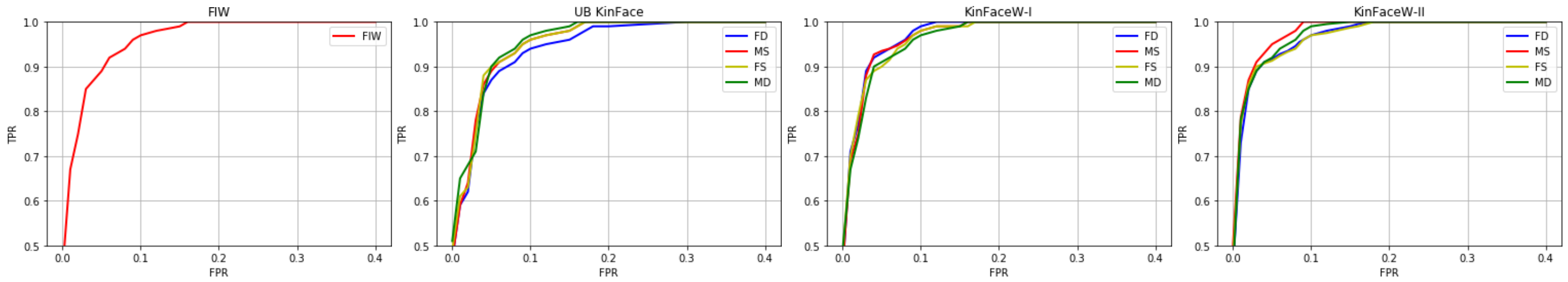

Fig. 5: The ROC curves of proposed models used VGG-19 as local feature extraction backbone. From left to right, these figures denote the results on the FIW dataset, UB KinFace dataset, KinFaceW-I dataset and KinFaceW-II dataset, respectively.

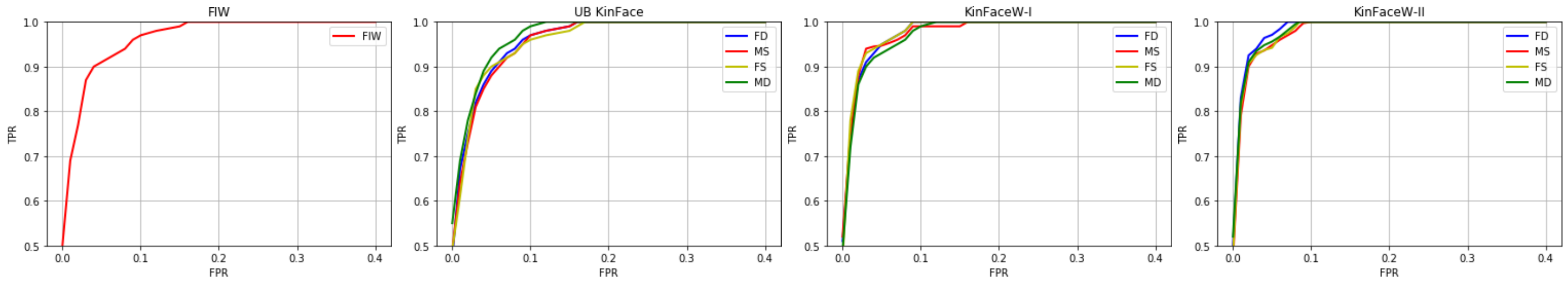

Fig. 6: The ROC curves of proposed models used resnet50 as local feature extraction backbone: from left to right is FIW dataset, UB KinFace dataset, KinFaceW-I dataset and KinFaceW-II dataset.

ferences between parent-offspring. Second, compared to the non-end-to-end methods (such as AdvKin [53], DTL [27] and ResNet+SDMLoss [56]), our method considers the feature extraction and similarity computation into one unified structure. It can promote the dependencies between feature learning and metrics calculation to obtain a more realistic and natural feature representation. Third, feature fusion can make better use of internal information, such that more discriminative information is exploited in different manners.

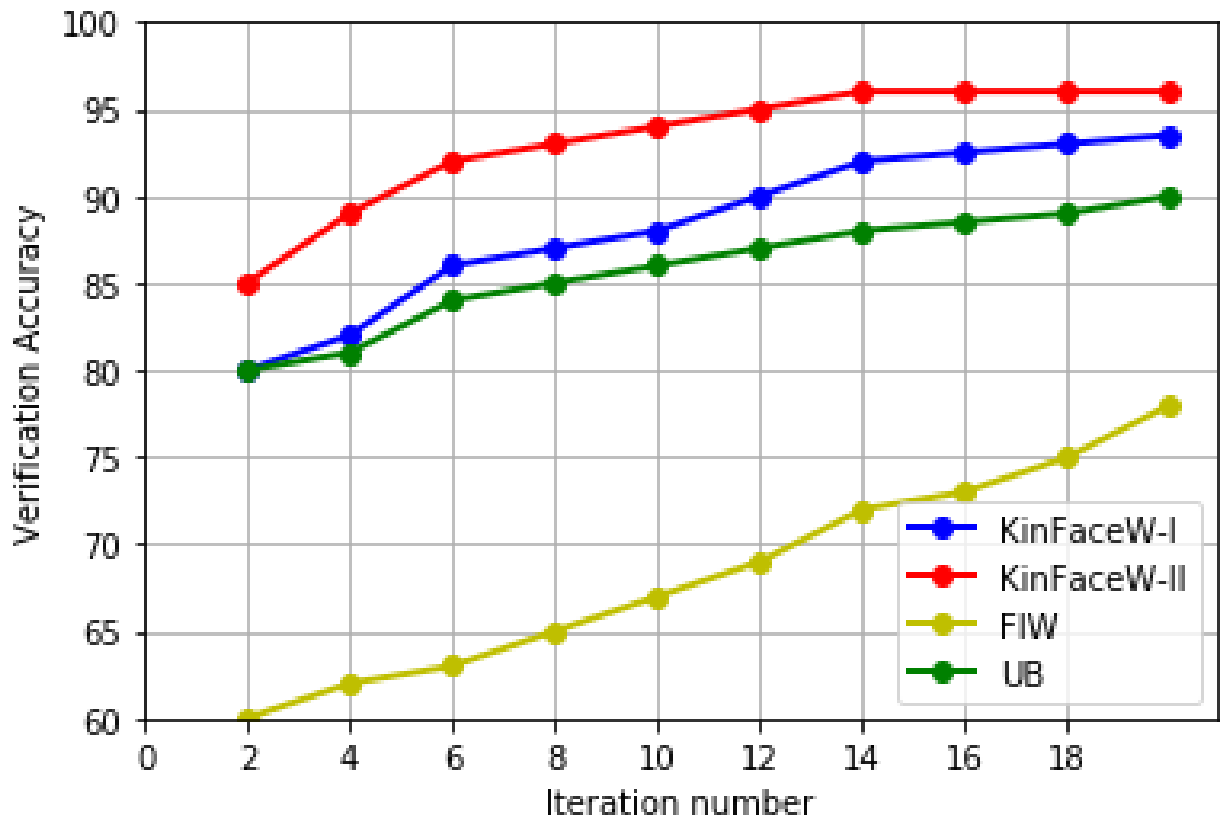

Fig. 7: Verification rate of CFIL versus different number of iteration on different kinship datasets.

TABLE V: Mean Verification Accuracies (%) of different model combinations on different datasets.

| Methods | F-S | F-D | M-S | M-D | overall [a] |
|---|---|---|---|---|---|
| FIW$_{Non-local}$ | 86.5 | 82.4 | 83.6 | 81.6 | 83 |
| FIW$_{local-VGG19}$ | 84.6 | 79.1 | 81.9 | 83.8 | 81.7 |
| FIW$_{local\ Resnet50}$ | 83.8 | 82.1 | 80.4 | 85.1 | 82 |
| FIW$_{Non\ local+local\ VGG19}$ | 91.6 | 87.5 | 88.8 | 89.4 | **88.6** |
| FIW$_{Non-local+local-Resnet50}$ | 91.8 | 89.3 | 89.7 | 90.2 | **89.45** |
| UB KinFace$_{Non-local}$ | 85 | 87.8 | 88 | 89 | 86.5 |
| UB KinFace$_{local-VGG19}$ | 85.2 | 84.6 | 84.7 | 86.3 | 84.9 |
| UB KinFace$_{local-Resnet50}$ | 82.6 | 87.5 | 87.2 | 89 | 85.3 |
| UB KinFace$_{Non-local+local-VGG19}$ | 88.4 | 89.3 | 90.7 | 91.2 | **88.9** |
| UB KinFace$_{Non-local+local-Resnet50}$ | 90.2 | 91.5 | 90.4 | 92 | **90.7** |
| KinFaceW-I$_{Non-local}$ | 90.6 | 92.3 | 92.5 | 91.1 | 92.3 |
| KinFaceW-I$_{local-VGG19}$ | 88.6 | 85.4 | 89.2 | 89.6 | 89.1 |
| KinFaceW-I$_{local-Resnet50}$ | 92.1 | 91.7 | 93.2 | 90.9 | 93.0 |
| KinFaceW-I$_{Non-local+local-VGG19}$ | 92.2 | 93.2 | 93.5 | 93.0 | **93.2** |
| KinFaceW-I$_{Non-local+local-Resnet50}$ | 97.5 | 95.6 | 96.8 | 94.3 | **97.2** |
| KinFaceW-II$_{Non-local}$ | 92.1 | 94.7 | 91.5 | 92.8 | 92.8 |
| KinFaceW-II$_{local-VGG19}$ | 88 | 86.9 | 91.2 | 91.7 | 89.5 |
| KinFaceW-II$_{local-Resnet50}$ | 93.7 | 91.9 | 94.2 | 91 | 92.7 |
| KinFaceW-II$_{Non-local+local-VGG19}$ | 92.8 | 93.0 | 94.3 | 93.8 | **93.5** |
| KinFaceW-II$_{Non-local+local-Resnet50}$ | 96.3 | 97.6 | 96.9 | 98.0 | **97.2** |

[a] overall: weighted accuracy.

*4) Ablation Study:* In this section, we conduct extensive ablation studies to verify the indispensability of different components in our CFIL framework. Typically, we try the experiments on different model combinations and the effectiveness of the non-local weighted operation compared to the fully-connected layer.

**Model Combinations** To verify the indispensability of interpolating the similarity calculations as the interior auxiliary weights into the deep CNN representation learning, we conduct ablation studies based on different models. We take the whole framework as two models, i.e., Non-Local Feature Extraction model and Local Feature Extraction model. We compare different backbones (VGG-19 [46] and Resnet-50 [47]) and different feature extraction methods on all datasets. As shown in Table V, a total of 20 combinations are verified on four datasets, and the best results have been labeled in boldface. All the training settings are the same as the experiments shown in the above section. The softmax classifier is used for performance prediction. The last channel is set to 2. From the experimental results, we find that the highest accuracy is up to 97.2% on the KinFaceW-II dataset when the backbone is Resnet50. Moreover, from Table V, we see that the full model is always superior to the degenerated models that remove any sub-module. This phenomenon further validates the indispensability and effectiveness of each sub-module in feature learning.

**Non-local Weighted Operation.** In this paper, we proposed a novel joint cross-generation feature interaction learning (CFIL) framework, which enhances the representation capability of learned features by performing a collaborative weighting strategy for kinship verification. Particularly, in non-local feature extraction model, because the conventional fully-connected layer can achieve similar results of mapping the distribution of features into deep latent space, we estimate the performance of the non-local weighted operation on non-local model by replacing it with a fully connected layer. Specifically, we compare different embeddings added to a shallow CNN model on all benchmark datasets. The comparison results in the FC column and non-local column of Table VI indicate that our non-local weighted operation can strengthen the response of the network at a more specific point level by capturing meaningful interactive information between different points.

## V. Conclusion

In this paper, we proposed a novel joint cross-generation feature interaction learning (CFIL) framework through a collaborative weighting strategy for feature extraction. This framework explored the characteristics of cross-generation relations by collaboratively extracting features from both parents-children image pairs. The proposed framework was built based on the non-local feature extraction module and the local feature extraction module.This method could excavate the multiple relationships between cross points so that more correlated information was preserved in the obtained features. In contrast to general face verification with separated models, the proposed method used a unified structure to integrate feature learning and similarity comparison, which effectively improved the dependencies between feature extraction and metric learning. Moreover, the similarity calculations were regarded as the interior auxiliary weights integrated into CNN representation learning. Extensive experiments demonstrated that our method could capture more natural and powerful interaction information for robust kinship verification. Furthermore, the superiority of the proposed method was substantially testified by comparing with some state-of-the-art algorithms.

Although our method has achieved the mentioned outstanding results, some issues still need to be addressed for practical applications. First, from the observation of datasets, the kinship pairs from KinFace-I and KinFace-II datasets are cropped from the same environmental conditions (such as same photos) [59], which is restricted and has no practical significance for actual recognition. Second, the existing datasets for kinship are not large enough to identify the real distributions and need to be shelved as far as possible. Therefore, we tend to use the FIW dataset to handle this task in the future.

Besides, due to the flexibility of our method, it not only can be used in kinship analysis but also for face verification, image segmentation, and video verification. It is worth noting that the weighted operation has strong abilities of differentiation and generalization and can be embedded in different models.

TABLE VI: Mean Verification Accuracies (%) of different components on different datasets.

| | FC | | | | | Non-local | | | | | Nothing [b] | | | | |
|---|---|---|---|---|---|---|---|---|---|---|---|---|---|---|---|
| Methods | F-S | F-D | M-S | M-D | overall | F-S | F-D | M-S | M-D | overall | F-S | F-D | M-S | M-D | overall [a] |
| FIW | 80.4 | 83.6 | 84.9 | 84.1 | 83.1 | 86.5 | 82.4 | 83.6 | 81.6 | 83 | 81.9 | 83.4 | 88.1 | 83.1 | 83.33 |
| UB KinFace | 85.2 | 81.6 | 83.7 | 87.4 | 83.3 | 85 | 87.8 | 88 | 89 | **86.5** | 84.28 | 85.3 | 80.5 | 83.4 | 83.7 |
| KinFaceW-I | 84.7 | 85 | 87.8 | 86.7 | 86.8 | 90.6 | 92.3 | 92.5 | 91.1 | **92.3** | 87.4 | 84 | 83.4 | 87.2 | 86.5 |
| KinFaceW-II | 86.8 | 84.2 | 85.6 | 88.6 | 86.3 | 92.1 | 94.7 | 91.5 | 92.8 | **92.8** | 88.6 | 89.1 | 87 | 88.9 | 88.4 |

[a] overall: weighted accuracy.
[b] : Nothing means no non-local weighted operation or FC layer.